\pdfoutput=1

\documentclass[11pt]{article}

\usepackage[]{ACL2023}

\usepackage{times}
\usepackage{bbm}
\usepackage{bm}
\usepackage{url}
\usepackage{enumerate}
\usepackage{amssymb}
\usepackage{makecell}
\usepackage{multirow}
\usepackage{pifont}
\usepackage{latexsym}
\usepackage{graphicx}
\usepackage{amsmath}
\usepackage{autobreak}
\usepackage{amsfonts}
\usepackage{subfigure}
\usepackage{booktabs}
\usepackage{colortbl}
\usepackage{algorithm}
\usepackage{algpseudocode}
\usepackage{color}
\DeclareMathOperator*{\argmax}{arg\,max}
\usepackage[T1]{fontenc}

\usepackage[utf8]{inputenc}

\usepackage{microtype}

\usepackage{inconsolata}

\title{Towards Understanding and Improving Knowledge Distillation \\for Neural Machine Translation}

\author{Songming Zhang, Yunlong Liang, Shuaibo Wang, \textbf{Yufeng Chen}\thanks{ \ \ Yufeng Chen is the corresponding author.}, \\
\textbf{Wenjuan Han}, \textbf{Jian Liu}, and 
\textbf{Jinan Xu}\\
Beijing Key Lab of Traffic Data Analysis and Mining, \\
Beijing Jiaotong University, Beijing, China \\
\texttt{\{smzhang22,yunlongliang,chenyf,wjhan,jianliu,jaxu\}@bjtu.edu.cn}}

\begin{document}
\maketitle
\begin{abstract}
Knowledge distillation (KD) is a promising technique for model compression in neural machine translation. 
However, where the knowledge hides in KD is still not clear, which may hinder the development of KD. 
In this work, we first unravel this mystery from an empirical perspective and show that the knowledge comes from the top-1 predictions of teachers, which also helps us build a potential connection between word- and sequence-level KD.
Further, we point out two inherent issues in vanilla word-level KD based on this finding. 
Firstly, the current objective of KD spreads its focus to whole distributions to learn the knowledge, yet lacks special treatment on the most crucial top-1 information.
Secondly, the knowledge is largely covered by the golden information due to the fact that most top-1 predictions of teachers overlap with ground-truth tokens, which further restricts the potential of KD.
To address these issues, we propose a novel method named \textbf{T}op-1 \textbf{I}nformation \textbf{E}nhanced \textbf{K}nowledge \textbf{D}istillation (TIE-KD). 
Specifically, we design a hierarchical ranking loss to enforce the learning of the top-1 information from the teacher. 
Additionally, we develop an iterative KD procedure to infuse more additional knowledge by distilling on the data without ground-truth targets. 
Experiments on WMT'14 English-German, WMT'14 English-French and WMT'16 English-Romanian demonstrate that our method can respectively boost Transformer$_{base}$ students by +1.04, +0.60 and +1.11 BLEU scores and significantly outperform the vanilla word-level KD baseline. 
Besides, our method shows higher generalizability on different teacher-student capacity gaps than existing KD techniques.
\end{abstract}

\section{Introduction} \label{sec:intro}
In recent years, neural machine translation (NMT) has made marvelous progress in generating high-quality translations \cite{rnnsearch,convs2s,transformer,liang-etal-2021-towards,liang-etal-2022-scheduled}, especially with some exquisite and deep model architectures \cite{wei-etal-2020-multiscale,li-etal-2020-shallow,liu-etal-2020-understanding,deepnet}. 
Despite their amazing performance on translation tasks, high computational and deployment costs still prevent these models from being applied in real life. 
On this problem, knowledge distillation (KD) \cite{liang2008structure,ori_kd,seq_kd,wu2020skip,chen-etal-2020-distilling,wang-etal-2021-selective,liang-etal-2021-modeling} is regarded as a promising solution for model compression, which aims to transfer the knowledge from these strong teacher models into compact student models. 

Generally, there are two categories of KD techniques, {\it i.e.}, word-level KD \cite{ori_kd,seq_kd,wang-etal-2021-selective} and sequence-level KD \cite{seq_kd}. 
(1) Word-level KD is conducted on each target token, where it shrinks the Kullback-Leibler (KL) divergence \cite{kullback1951information} between the predicted distributions from the student and the soft targets from the teacher.
In these soft targets, the knowledge was previously deemed to come from the probability relationship between negative candidates (\textit{i.e.}, the correlation information) \cite{ori_kd,understanding_kd,jafari-etal-2021-annealing}. 
(2) Sequence-level KD instead requires no soft target and directly encourages students to maximize the sequence probability of the final translation decoded by the teacher. 
Although both techniques work quite differently, they still achieve similarly superior effectiveness. 
Therefore, we raise two heuristic questions on KD in NMT: 
\begin{itemize}
    \item \textit{Q1: Where does the knowledge actually come from during KD in NMT?}
    \item \textit{Q2: Is there any connection between the word- and the sequence-level KD techniques?}
\end{itemize}

To answer these two questions, we conduct an empirical study that starts from word-level KD to find out where the knowledge hides in the teacher's soft targets and then explore whether the result can be expanded to sequence-level KD. 
As a result, we summarize several intriguing findings:
\begin{enumerate}[i.]
    \item Compared to the correlation information, the information of the teacher's top-1 predictions (\textit{i.e.}, the top-1 information) actually determines the benefit of word-level KD (\S\ref{sec:determine}).
    \item The correlation information can be successfully learned by students during KD but fails to improve their final performance (\S\ref{sec:really_learn}).
    \item Extending the top-1 information to top-$k$ does not lead to further improvement (\S\ref{sec:topk_info}).
    \item The top-1 information is important even when the teacher is under-confident in its top-1 predictions (\S\ref{sec:diff_top1_conf}).
    \item Similar importance of the top-1 information can also be verified on sequence-level KD (\S\ref{sec:expand_seq}).
\end{enumerate}
These findings sufficiently prove that \textbf{1) the knowledge actually comes from the top-1 information of the teacher during KD in NMT}, and \textbf{2) the two kinds of KD techniques can be connected from the perspective of the top-1 information}.

On these grounds, we further point out that there are two inherent issues in vanilla word-level KD.
Firstly, as the source of teachers' knowledge, the top-1 information receives no special treatment in the training objective of vanilla word-level KD since the KL divergence directly optimizes the entire distribution.
Secondly, since most top-1 predictions of strong teachers overlap with ground-truth tokens (see the first row of Tab.\ref{tab:overlap}), the additional knowledge from teachers beyond the golden information is poor and the potential of word-level KD is largely limited (see the second row of Tab.\ref{tab:overlap}). 
To address these issues, we propose a new KD method named \textbf{T}op-1 \textbf{I}nformation \textbf{E}nhanced \textbf{K}nowledge \textbf{D}istillation (TIE-KD) for NMT. 
Specifically, we first design a hierarchical ranking loss that can enforce the student model to learn the top-1 information through ranking the top-1 predictions of the teacher as its own top-1 predictions.
Moreover, we develop an iterative KD procedure to expose more input data without ground-truth targets for KD to exploit more knowledge from the teacher.
\begin{table}[]
    \centering
    \resizebox{\linewidth}{!}{
    \begin{tabular}{c|c|c|c}
        \bottomrule
        Datasets & En-De & En-Fr & En-Ro \\
        \hline
        Top-1 Overlap Rate & 68\% & 78\% & 94\% \\
        \hline
        $\Delta$ from Word-level KD & +0.61 & +0.13 & +0.18 \\
        \toprule
    \end{tabular}
    }
    \caption{The overlap rates between the top-1 predictions of teachers and ground-truth tokens on WMT'14 English-German (En-De), WMT'14 English-French (En-Fr) and WMT'16 English-Romanian (En-Ro) and the corresponding improvement ($\Delta$) of BLEU scores brought by word-level KD on the test set of these tasks\protect\footnotemark.}
    \label{tab:overlap}
\end{table}
\footnotetext{We random sample 3000 target sentences in the training set of each task to calculate the approximate overlap rates.}

We evaluate our TIE-KD method on three WMT benchmarks, {\it i.e.}, WMT'14 English-German (En-De), WMT'14 English-French (En-Fr) and WMT'16 English-Romanian (En-Ro). 
Experimental results show that our method can boost Transformer$_{base}$ students by +1.04, +0.60, +1.11 BLEU scores and significantly outperforms the vanilla word-level KD approach.
Besides, we test the performance of existing KD techniques in NMT and our TIE-KD under different teacher-student capacity gaps and show the stronger generalizability of our method on various gaps.

Our contributions are summarized as follows\footnote{The code is publicly available at: \url{https://github.com/songmzhang/NMT-KD}.}:
\begin{itemize}
    \item To the best of our knowledge, we are the first to explore where the knowledge hides in KD for NMT and unveil that it comes from the top-1 information of the teacher, which also helps us build a connection between word- and sequence-level KD.
    \item Further, we point two issues in vanilla word-level KD and propose a novel KD method named Top-1 Information Enhanced Knowledge Distillation (TIE-KD) to address them. Experiments on three WMT benchmarks demonstrate its effectiveness and superiority.
    \item  We investigate the effects of current KD techniques in NMT under different teacher-student capacity gaps and show the stronger generalizability of our approach to various gaps.
\end{itemize}
\begin{figure*}[t]
    \centering
    \subfigure[vanilla word-level KD]{
        \includegraphics[width=0.19\linewidth]{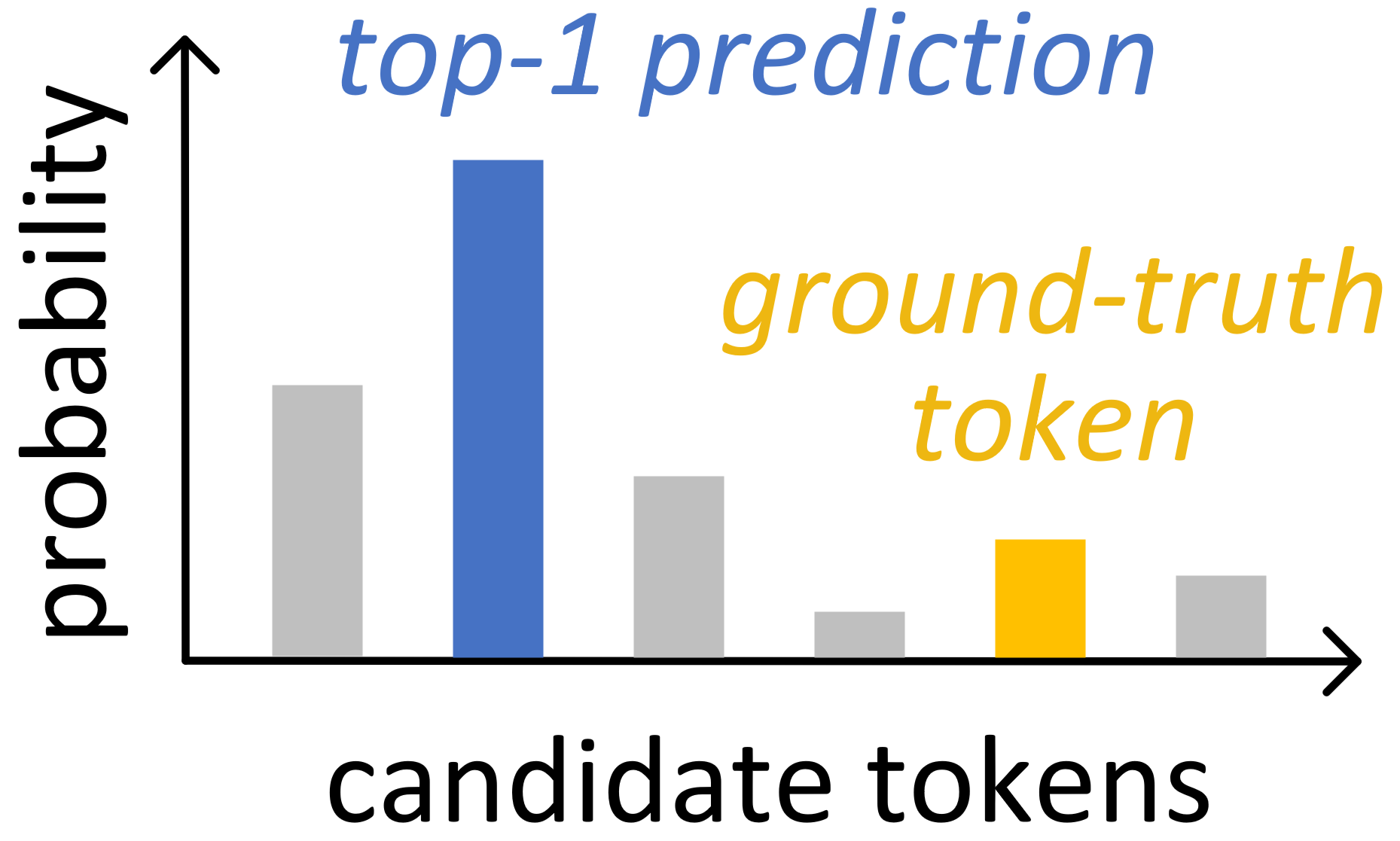}
    }
    \quad\quad
    \subfigure[\textit{w/o} correlation info]{
        \includegraphics[width=0.19\linewidth]{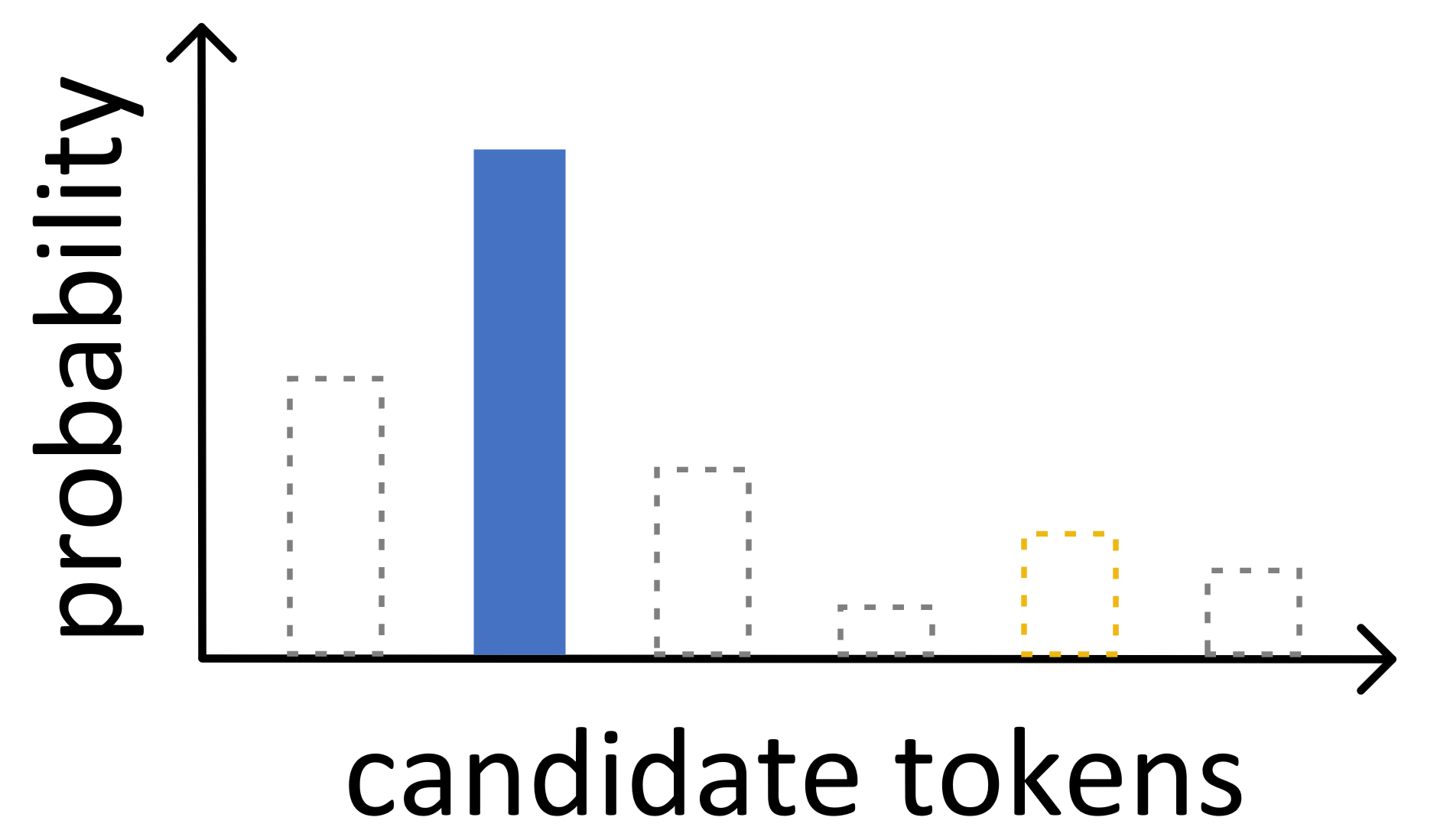}
    }
    \quad\quad
    \subfigure[\textit{w/o} top-1 info]{
        \includegraphics[width=0.19\linewidth]{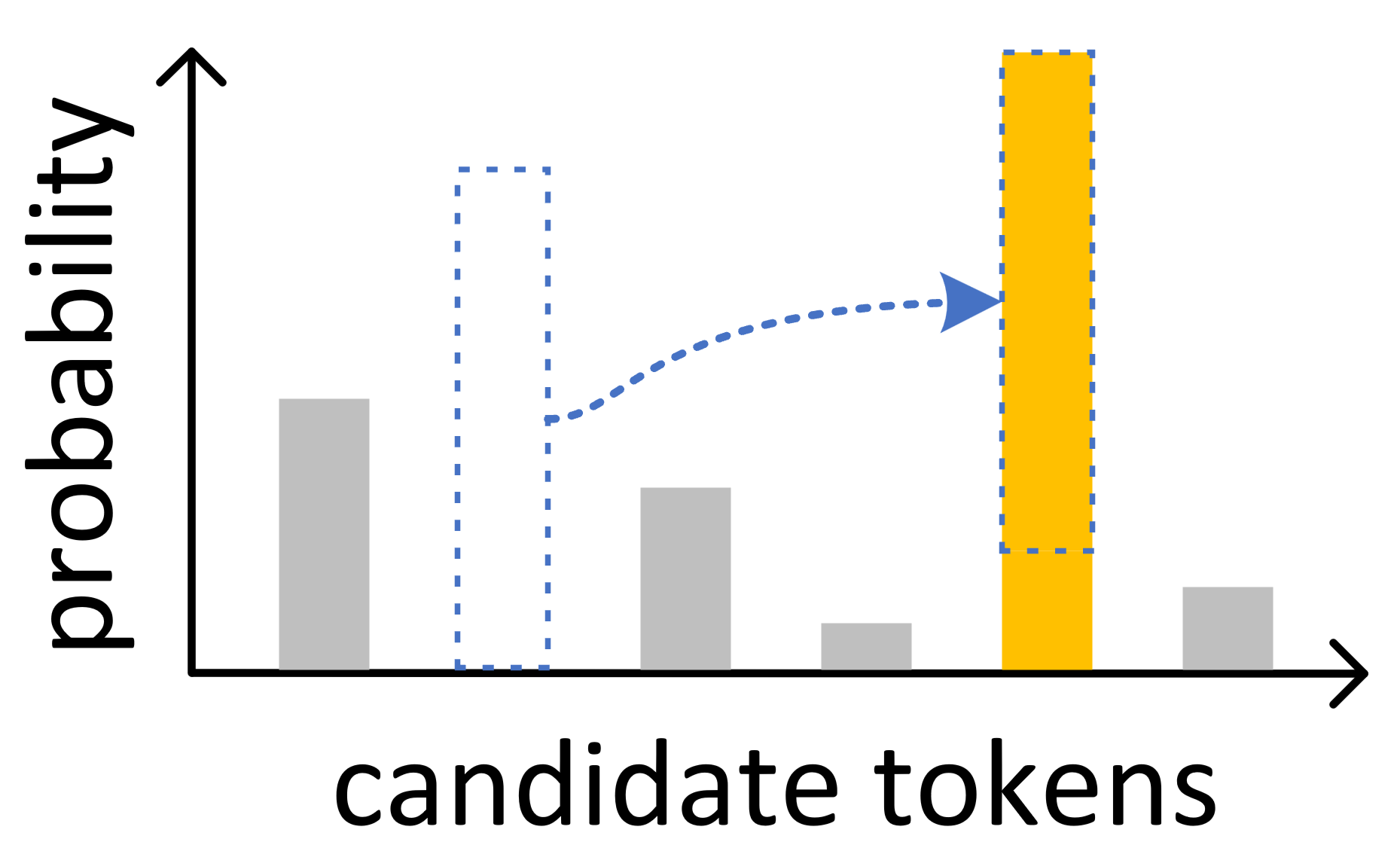}
    }
    \quad\quad
    \subfigure[\textit{w/o} KD]{
        \includegraphics[width=0.19\linewidth]{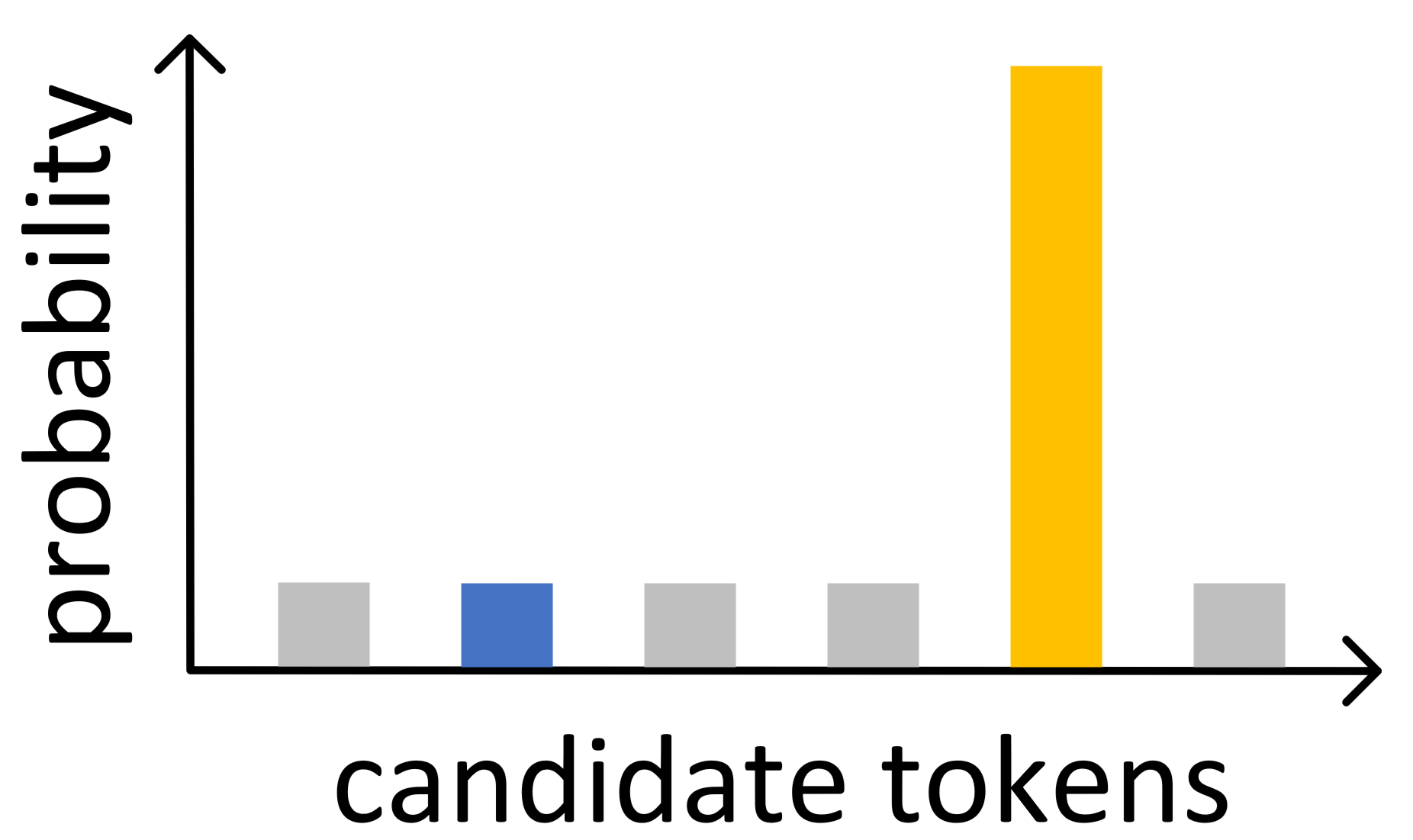}
    }
    
    \caption{Removing different information from the original soft targets provided by the teacher during word-level KD. Note that the soft target in ``\textit{w/o} KD'' is equivalent to the soft target of label smoothing.}
    \label{fig:probe}
\end{figure*}
\section{Background}
\subsection{Neural Machine Translation}


Given a source sentence with $M$ tokens $\mathbf{x}=\{x_1, x_2, \ldots, x_M\}$ and the corresponding target sentence with $N$ tokens $\mathbf{y}=\{y_1, y_2, \ldots, y_N\}$, NMT models are trained to maximize the probability of each target token conditioning on the source sentence by the cross-entropy (CE) loss:
\begin{equation} \label{eq:ce_loss}
     \mathcal{L}_{ce}= -\sum_{j=1}^N\log p(y_j^{\ast}|\mathbf{y}_{<j},\mathbf{x};\theta),
\end{equation}
where $y_j^{\ast}$ and $\mathbf{y}_{<j}$ denote the ground-truth target and the target-side previous context at time step $j$, respectively. And $\theta$ is the model parameter.
\subsection{Word-level Knowledge Distillation}
Word-level KD \cite{seq_kd} aims to minimize the KL divergence between the output distributions of the teacher model and the student model on each target token. Formally, given the probability distribution $q(\cdot)$ from the teacher model, the KL divergence-based loss is formulated as follows:
\begin{flalign} \label{eq:kl_loss}
&\mathcal{L}_{kd}=\mathcal{L}_{\rm KL}=&\\
& \ \ \sum_{j=1}^N D_{\rm KL}\Big(q(y_j|\mathbf{y}_{<j},\mathbf{x};\theta_t)\big|\big|p(y_j|\mathbf{y}_{<j},\mathbf{x};\theta_s)\Big),& \nonumber             
\end{flalign}
where $\theta_t$ and $\theta_s$ denote the model parameters of the teacher and the student, respectively.

Then, the overall loss function of word-level KD is the linear interpolation between the CE loss and the KL divergence loss:
\begin{equation} \label{eq:ori_final_loss}
    \mathcal{L}_{word\raisebox{0mm}{-}kd}=(1-\alpha)\mathcal{L}_{ce} + \alpha\mathcal{L}_{kd}.
\end{equation}
\subsection{Sequence-level Knowledge Distillation}
Sequence-level KD \cite{seq_kd} encourages the student model to imitate the sequence probabilities of the translations from the teacher model. 
To this end, it optimizes the student model through the following approximation:
\begin{align} \label{eq:seq_kd}
    \mathcal{L}_{seq\raisebox{0mm}{-}kd}&=-\sum_{\mathbf{y}\in\mathcal{Y}}Q(\mathbf{y}|\mathbf{x};\theta_t)\log P(\mathbf{y}|\mathbf{x};\theta_s) \nonumber \\
    &\approx-\log P(\widehat{\mathbf{y}}|\mathbf{x};\theta_s),
\end{align}
where $\mathcal{Y}$ denotes the hypothesis space of the teacher and $\widehat{\mathbf{y}}$ is the approximate result through the teacher's beam search.

\section{Probing the Knowledge of KD in NMT}
\label{sec:sec3}
In this section, we start from word-level KD and offer exhaustive empirical analyses on 1) the determining information in word-level KD (\S\ref{sec:determine}); 2) whether the correlation information has been learned (\S\ref{sec:really_learn}); 3) whether there are more benefits when extending the top-1 to top-$k$ information (\S\ref{sec:topk_info}) and 4) the importance of the top-1 information on soft targets with different confidence (\S\ref{sec:diff_top1_conf}). 
Then we expand the conclusion to sequence-level KD (\S\ref{sec:expand_seq}) and lastly revisit KD for NMT from a novel view (\S\ref{sec:rethink}).
\subsection{Which Information Determines the Performance of Word-level KD?} 
\label{sec:determine}
In word-level KD, the relative probabilities between negative candidates in the soft targets from the teacher contain rich correlation information, which is previously deemed to carry knowledge from the teacher \cite{ori_kd,understanding_kd,jafari-etal-2021-annealing}. 
However, in practice, strong teachers usually have high confidence in their top-1 predictions while retaining little probability mass for other candidates. 
Hence, to study the mystery of KD, it is necessary to first investigate the real effects of the correlation information and the top-1 prediction information during KD and then figure out which one actually determines the performance of KD.

\begin{table}[]
    \centering
    \resizebox{\linewidth}{!}{
    \begin{tabular}{l|l|c|c}
        \bottomrule
        \textbf{Task} & \textbf{Model} & \textbf{TA} & \textbf{BLEU} \\
        \hline
        \multirow{4}*{En-De} & (a) vanilla word-level KD & \cellcolor{blue!25}88.98 & \cellcolor{cyan!25}26.66 \\
        ~ & (b) \textit{w/o} correlation info & \cellcolor{blue!20}88.69 & \cellcolor{cyan!35}26.76 \\
        ~ & (c) \textit{w/o} top-1 info & \cellcolor{blue!10}87.49 & \cellcolor{cyan!10}26.43 \\
        ~ & (d) \textit{w/o} KD & \cellcolor{blue!5}87.22 & \cellcolor{cyan!5}26.37 \\
        \hline
        \multirow{4}*{En-Fr} & (a) vanilla word-level KD & \cellcolor{blue!25}89.31 & \cellcolor{cyan!25}34.94 \\
        ~ & (b) \textit{w/o} correlation info & \cellcolor{blue!20}89.19 & \cellcolor{cyan!35}35.09 \\
        ~ & (c) \textit{w/o} top-1 info & \cellcolor{blue!10}88.34 & \cellcolor{cyan!10}34.33 \\
        ~ & (d) \textit{w/o} KD & \cellcolor{blue!5}88.33 & \cellcolor{cyan!5}34.69 \\
        \hline
        \multirow{4}*{En-Ro} & (a) vanilla word-level KD & \cellcolor{blue!20}83.98 & \cellcolor{cyan!25}34.29 \\
        ~ & (b) \textit{w/o} correlation info & \cellcolor{blue!25}84.27 & \cellcolor{cyan!35}34.30 \\
        ~ & (c) \textit{w/o} top-1 info & \cellcolor{blue!10}83.73 & \cellcolor{cyan!10}34.02 \\
        ~ & (d) \textit{w/o} KD & \cellcolor{blue!5}83.34 & \cellcolor{cyan!5}34.04 \\
        \toprule
    \end{tabular}
    }
    \caption{Top-1 Agreement rates (\%) and BLEU scores (\%) of different soft targets during KD on the validation sets of the three tasks. Deeper colors represent better performance on the corresponding metrics.}
    \label{tab:remove_info}
\end{table}

To this end, during word-level KD, we separately remove the top-1 information and the correlation information from the original soft targets of the teacher (as depicted in Fig.\ref{fig:probe}) and then observe the corresponding performance.
Besides the BLEU score, we also introduce a new metric, namely the \textbf{Top-1 Agreement (TA)} rate, which calculates the overlap rate of the top-1 predictions between the student and the teacher on each position under the teacher-forcing mode.
As shown in Tab.\ref{tab:remove_info}, the performance slightly increases when we remove the probabilities of all other candidates except for the top-1 ones in soft targets (see Fig.\ref{fig:probe}(b))\footnote{Considering the regularization effect, we do not add a uniform distribution to complement the removed probability. Please refer to Appendix \ref{sec:whynot_renorm} for more detailed explanations.}. 
However, when we only remove the top-1 information and keep the remaining correlation information (see Fig.\ref{fig:probe}(c))\footnote{Note that we do not simply remove the probability of the top-1 prediction, but add this probability to the ground-truth token to maintain the correctness of the distribution, \textit{i.e.}, the soft target is unchanged if its top-1 prediction is correct.}, the performance of KD drops close to the baseline without any KD. 
Moreover, we observe that the TA rates are well correlated with the final BLEU scores among these students.
Therefore, we conjecture that the top-1 information is the one that actually determines the performance of word-level KD (answer to \emph{Q1}). 

\subsection{Can Student Models Really Learn the Correlation Information?}\label{sec:really_learn}
To further confirm the above conjecture, we examine whether the student models have successfully learned the correlation information of the teacher during KD. 
To achieve this, we design two metrics to measure the ranking similarities between token rankings from the student and the teacher, named top-$k$ edit distance and top-$k$ ranking distance.
\paragraph{Top-$\bm{k}$ Edit Distance.} Given the top-$k$ predictions of the teacher at time step $j$ as $[y_j^{t_1},...,y_j^{t_k}]$ and the ones of the student as $[y_j^{s_1},...,y_j^{s_k}]$, the top-$k$ edit distance can be expressed as:
\begin{equation}
    \mathcal{D}_{edit}=\frac{1}{N}\sum_j^Nf([y_j^{t_1},...,y_j^{t_k}], [y_j^{s_1},...,y_j^{s_k}]), \nonumber
\end{equation}
where $f(\cdot, \cdot)$ calculates the edit distance.
\paragraph{Top-$\bm{k}$ Ranking Distance.} For each $y_j^{t_i}$ in $[y_j^{t_1},...,y_j^{t_k}]$, this metric measures the average ranking distance between its original rank $i$ from the teacher, and the corresponding rank from the student, denoted as $r_s(y_j^{t_i})$:
\begin{equation}
    \mathcal{D}_{rank}=\frac{1}{Nk}\sum_j^N\sum_i^k\min(k, |i-r_s(y_j^{t_i})|). \nonumber
\end{equation}

\begin{table}[]
    \centering
    \resizebox{\linewidth}{!}{
    \begin{tabular}{l|l|c|c|c}
        \bottomrule
        \textbf{Task} & \textbf{Model} & \textbf{$\bm{\mathcal{D}_{edit}}\downarrow$} & \textbf{$\bm{\mathcal{D}_{rank}}\downarrow$} & \textbf{BLEU} \\
        \hline
        \multirow{4}*{En-De} & (a) vanilla Word-KD & \cellcolor{pink!60}2.506 & \cellcolor{pink!60}1.571 & \cellcolor{cyan!20}26.66 \\
        ~ & (b) \textit{w/o} correlation info & \cellcolor{pink!20}2.697 & \cellcolor{pink!20}1.791 & \cellcolor{cyan!40}26.76 \\
        ~ & (c) \textit{w/o} top-1 info & \cellcolor{pink!40}2.601 & \cellcolor{pink!40}1.656 & \cellcolor{cyan!10}26.43 \\
        ~ & (d) \textit{w/o} KD & \cellcolor{pink!5}2.739 & \cellcolor{pink!5}1.820 & \cellcolor{cyan!5}26.37 \\
        \hline
        \multirow{4}*{En-Fr} & (a) vanilla Word-KD & \cellcolor{pink!40}2.515 & \cellcolor{pink!40}1.588 & \cellcolor{cyan!20}34.94 \\
        ~ & (b) \textit{w/o} correlation info & \cellcolor{pink!5}2.616 & \cellcolor{pink!5}1.696 & \cellcolor{cyan!40}35.09 \\
        ~ & (c) \textit{w/o} top-1 info & \cellcolor{pink!60}2.495 & \cellcolor{pink!60}1.563 & \cellcolor{cyan!5}34.33 \\
        ~ & (d) \textit{w/o} KD & \cellcolor{pink!20}2.587 & \cellcolor{pink!20}1.657 & \cellcolor{cyan!10}34.69 \\
        \hline
        \multirow{4}*{En-Ro} & (a) vanilla Word-KD & \cellcolor{pink!40}2.915 & \cellcolor{pink!40}2.000 & \cellcolor{cyan!20}34.29 \\
        ~ & (b) \textit{w/o} correlation info & \cellcolor{pink!5}3.025 & \cellcolor{pink!5}2.138 & \cellcolor{cyan!40}34.30 \\
        ~ & (c) \textit{w/o} top-1 info & \cellcolor{pink!60}2.893 & \cellcolor{pink!60}1.998 & \cellcolor{cyan!5}34.02 \\
        ~ & (d) \textit{w/o} KD & \cellcolor{pink!20}2.967 & \cellcolor{pink!20}2.083 & \cellcolor{cyan!10}34.04 \\
        \toprule
    \end{tabular}
    }
    \caption{Ranking similarities between the students and the teachers and the corresponding BLEU scores (\%)\protect\footnotemark.}
    \label{tab:rank_compare}
\end{table}
\footnotetext{Here we set $k$ to 5 for both $\mathcal{D}_{edit}$ and $\mathcal{D}_{rank}$ since different $k$ does not change the conclusion in our experiments.}

We compare the students above based on these two metrics and list the results in Tab.\ref{tab:rank_compare}. 
Clearly, the students perform better on both $\mathcal{D}_{edit}$ and $\mathcal{D}_{rank}$ when the soft targets contain correlation information ((a),(c) \textit{vs}. (b),(d)), indicating that student models can successfully learn the correlation information from the teacher. 
However, this ranking performance fails to bring better performance of KD, as measured by BLEU scores.
Thus, these results negate the previous perception that the correlation information carries the knowledge during KD, which also supports our conjecture in Sec.\ref{sec:determine}.

\subsection{Does Knowledge Increase with Top-$\bm{k}$ Information?} \label{sec:topk_info}
As the importance of the top-1 information for transferring knowledge in word-level KD has been validated, we further investigate whether more knowledge can be exploited by extending top-1 information to top-$k$ information\footnote{Equivalent to vanilla word-level KD when $k=|V|$.}.
Similar to Fig.\ref{fig:probe}(b), we keep the top-$k$ probabilities in the original soft target and remove others to extract its top-$k$ information.
However, the results in Tab.\ref{tab:topk_info} give a negative answer that more information does not bring significantly more knowledge.
Thus, we can believe that almost all the knowledge of the teacher in word-level KD comes from the teacher's top-1 information, even though the whole distribution is distilled to the student. 
\begin{table}[]
    \centering
    \resizebox{\linewidth}{!}{
    \begin{tabular}{c|l|ccccc}
        \bottomrule
        \multicolumn{2}{c|}{$k$} & 1 & 3 & 5 & 30 & $|V|$ \\
        \hline
        \multirow{3}{*}{BLEU} & En-De & 26.76 & 26.74 & 26.76 & 26.70 & 26.66 \\
        \cline{2-7}
        ~ & En-Fr & 35.09 & 34.91 & 34.79 & 34.79 & 34.94 \\
        \cline{2-7}
        ~ & En-Ro & 34.30 & 34.38 & 34.28 & 34.30 & 34.29 \\
        \toprule
    \end{tabular}
    }
    \caption{BLEU scores (\%) of word-level KD with top-$k$ information on the validation set of the three tasks. $|V|$ is the vocabulary size.}
    \label{tab:topk_info}
\end{table}
\subsection{Does Top-1 Information Work in All Soft Targets?} \label{sec:diff_top1_conf}
\begin{figure}
    \centering
    \includegraphics[width=\linewidth]{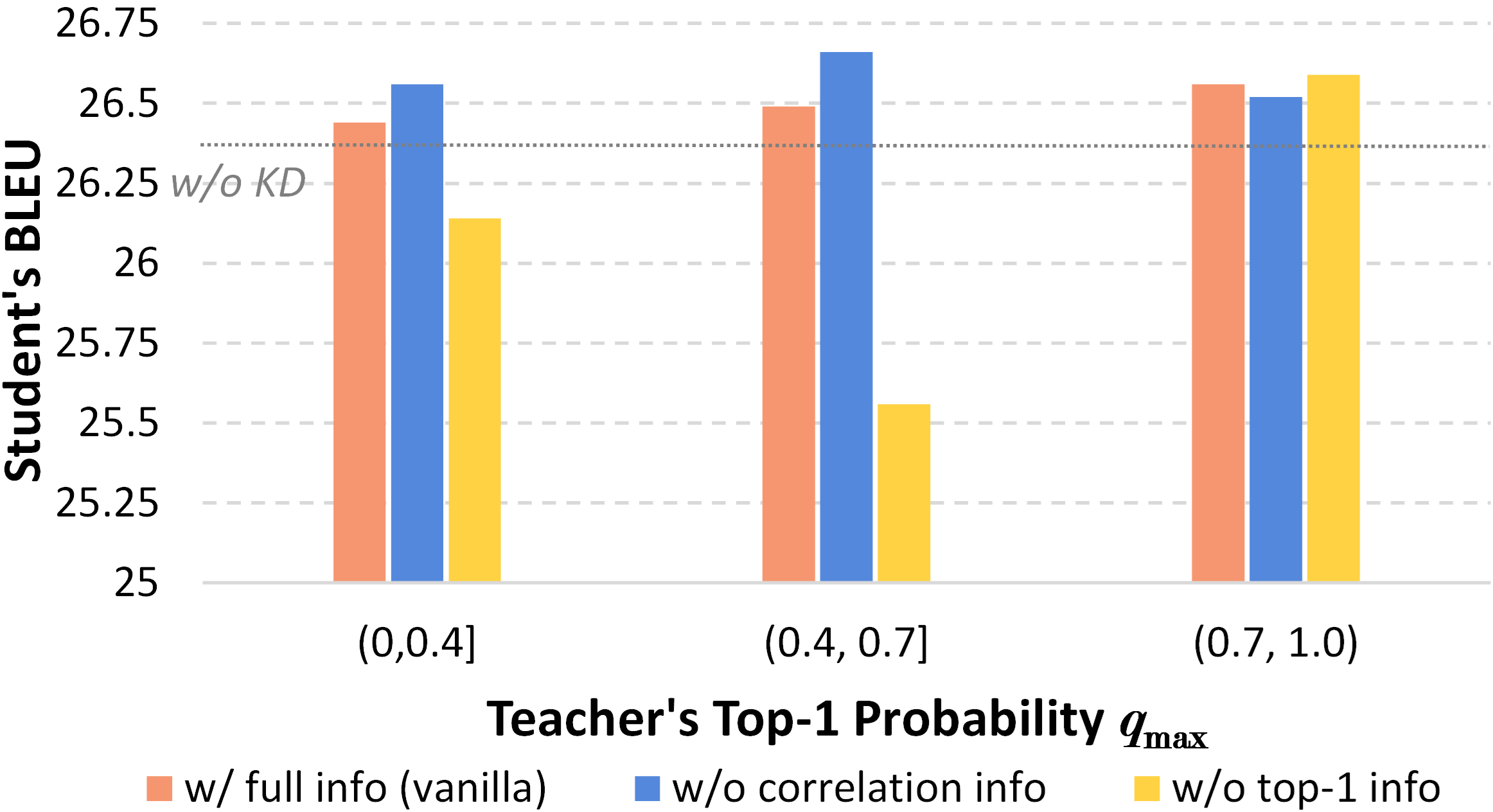}
    \caption{BLEU scores (\%) of KD with different information in three intervals of soft targets on the validation set of the WMT'14 En-De task.}
    \label{fig:interval}
\end{figure}
Although the previous results have coarsely located the knowledge in word-level KD on the top-1 information of the teacher, it is still not clear whether this holds for all types of soft targets, especially when the teacher is under-confident in its top-1 predictions. 
Towards this end, we divide the soft targets of the teacher into three intervals \cite{wang-etal-2021-selective} based on their top-1 probabilities: $(0.0, 0.4]$, $(0.4, 0.7]$, and $(0.7, 1.0)$. 
Then we separately conduct the same KD processes as described in Fig.\ref{fig:probe}, only using the soft targets in one of these intervals. 
Surprisingly, the results in Fig.\ref{fig:interval} show that even when the teacher is not so confident (\textit{i.e.}, $q_{\rm max}\leq0.7$) in its top-1 predictions, using only the top-1 information (\textit{i.e.}, the blue bars) still achieves better performance than using the full information in corresponding soft targets. 
However, in these cases, removing the top-1 information in soft targets largely degrades the performance of the students.
We conjecture that these under-confident top-1 predictions of the teacher can serve as hints for students to learn the difficult ground-truth labels, while the correlation information in these cases carries more noise than real knowledge for students.

\begin{table}[]
    \centering
    \resizebox{\linewidth}{!}{
    \begin{tabular}{c|c|c|c}
        \bottomrule
        \textbf{ID} & \textbf{top-1} ($\approx$70\%) & \textbf{non-top-1} ($\approx$30\%) & \textbf{BLEU} \\
        \hline
        1 & \checkmark & \checkmark & 26.86 \\
        2 & \checkmark & \ding{55} & 26.83 \\
        3 & \ding{55} & \checkmark & 2.36 \\
        4 & \checkmark (use fixed 30\%) & \ding{55} & 26.06 \\
        5 & \checkmark & \checkmark + word-level top-1 info & \textbf{26.96} \\
        \toprule
    \end{tabular}
    }
    \caption{BLEU scores (\%) of sequence-level KD on the validation set of the WMT'14 En-De task when we separately use the top-1 and the non-top-1 targets of the teacher in the teacher's translations during KD.}
    \label{tab:seqkd}
\end{table}
\subsection{Expanding to Sequence-level KD} \label{sec:expand_seq}
Inspired by the analyses on word-level KD, we move on to sequence-level KD and decompose its loss function in Eq.\eqref{eq:seq_kd} into a word-level form:
\begin{align} \label{eq:decompose}
    \mathcal{L}_{seq\raisebox{0mm}{-}kd}&\approx-\log P(\widehat{\mathbf{y}}|\mathbf{x};\theta_s) \nonumber \\
    &=-\sum^N_j \log p(\widehat{y}_j|\widehat{\mathbf{y}}_{<j},\mathbf{x};\theta_s),
\end{align}
where $\widehat{y}_j$ is the teacher-decoded target for students at time step $j$.
Considering the similar word-level form, it is intuitive to speculate that the top-1 information may also matter in sequence-level KD.
To verify this, we divide the targets $\widehat{y}_j$ into the top-1 and the non-top-1 predictions of the teacher\footnote{There are about 70\% top-1 predictions and 30\% non-top-1 predictions selected by the teacher's beam search during decoding.} and investigate the respective effects of these targets by separately using them during sequence-level KD.
As shown in Tab.\ref{tab:seqkd}, there is only a negligible performance change when we only use the top-1 targets for KD (row 1 \textit{vs}. row 2). 
However, if we only use the non-top-1 targets, the BLEU score drastically drops (row 1 \textit{vs}. row 3).
Moreover, considering the different proportions of the two kinds of targets in the teacher's translations (\textit{i.e.},70\% \textit{vs}. 30\%), we also use a fixed part (the same amount as the non-top-1 targets) of the top-1 targets for a fair comparison, and the performance is still steady (row 2 \textit{vs}. row 4) and much better than using only the non-top-1 targets (row 3 \textit{vs}. row 4).
Interestingly, by adding additional word-level top-1 information to the non-top-1 part, the performance of sequence-level KD further improves (row 1 \textit{vs}. row 5).
Therefore, we can also confirm the importance of the top-1 information in sequence-level KD.

\subsection{Rethinking KD in NMT from the Perspective of the Top-1 Information} \label{sec:rethink}
Through the above analyses, we verify the importance of the teacher's top-1 information on both KD techniques, which actually reflects a potential connection between them. 
A brief theoretical analysis on this connection is provided in Appendix \ref{sec:theoretical}.
In short, the two kinds of techniques share a unified objective that imparts the teachers' top-1 predictions to student models at each time step.
Thus, we believe that they are well connected on their similar working mechanisms (answer to \emph{Q2}).

Further, we revisit word-level KD from this perspective and find two inherent issues.
Firstly, the KL divergence-based objective in vanilla word-level KD directly optimizes whole distributions of students, while lacking specialized learning of the most important top-1 information.
Secondly, since the top-1 predictions of the teacher mostly overlap with the ground-truth targets, the knowledge from the teacher is largely covered by the ground-truth information, which largely limits the potential of word-level KD.
Therefore, we claim that the performance of the current word-level KD approach is far from perfect and the solutions to these problems are urgently needed.

\section{Top-1 Information Enhanced Knowledge Distillation for NMT} \label{sec:method}

To address the aforementioned issues in word-level KD, in this section, we introduce our method named \textbf{T}op-1 \textbf{I}nformation \textbf{E}nhanced \textbf{K}nowledge \textbf{D}istillation (TIE-KD), which includes a hierarchical ranking loss to boost the learning of the top-1 information from the teacher (\S\ref{sec:hrl}) and an iterative knowledge distillation procedure to exploit more knowledge from the teacher (\S\ref{sec:iter_kd}).

\subsection{Hierarchical Ranking Loss} \label{sec:hrl}
To help student models better grasp the top-1 information during distillation, we design a new loss named hierarchical ranking loss. 
To gently achieve this goal, we first encourage the student to rank the teacher's top-$k$ predictions as its own top-$k$ predictions and then rank the teacher's top-1 prediction over these top-$k$ predictions. 
Formally, given the student's top-$k$ predictions as $[y_j^{s_1},...,y_j^{s_k}]$ and the teacher's top-$k$ predictions as $[y_j^{t_1},...,y_j^{t_k}]$, the hierarchical ranking loss $\mathcal{L}_{hr}$ can be expressed as:
\begin{equation}
    \label{eq:hr_loss}
    \begin{split}
        \mathcal{L}_{hr}&=\sum_j^N\Big(\sum_u^k\sum_v^k\max\big\{0,  \\
        &\mathbbm{1}\{q(y_j^{t_u})>q(y_j^{s_v})\}(p(y_j^{s_v})-p(y_j^{t_u}))\big\} \\
        &+\sum_u^k\max\big\{0, p(y_j^{t_u})-p(y_j^{t_1})\big\}\Big),
    \end{split}
\end{equation}
where $p(\cdot)$ and $q(\cdot)$ are the probabilities from the student model and the teacher model, respectively. And $\mathbbm{1}\{\cdot\}$ is an indicator function. 

In this way, the student model can be enforced to rank the top-1 predictions of the teacher to its own top-1 places, and thus it can explicitly enhance the learning of the knowledge from the teacher. 
Then, we add this loss to the original KL divergence loss, \textit{i.e.}, Eq.\eqref{eq:kl_loss}, forming a new loss for KD:
\begin{equation} \label{eq:new_kd_loss}
    \mathcal{L}_{kd}=\mathcal{L}_{\rm KL}+\mathcal{L}_{hr}.
\end{equation}

\subsection{Iterative Knowledge Distillation}
\label{sec:iter_kd}
\begin{algorithm}[t]
    \caption{Iterative Knowledge Distillation}
    \label{algo:iter_kd}
    \begin{algorithmic}[1] 
        \Require source and target data in current mini-batch $(\mathbf{x},\mathbf{y})$; student model $\mathcal{S}$; teacher model $\mathcal{T}$; iteration times $N$;
        \State Initialize $\mathbf{y}^0=\mathbf{y}$; $\mathcal{L}_{kd}=0$;
        \State Compute $\mathcal{L}_{ce}$ based on Eq.\eqref{eq:ce_loss}
        \For{$i$ in $1,2,...,N$}
        \State $p^i=\mathcal{S}(\mathbf{x};\mathbf{y}^{i-1})$ \Comment{\textit{probability distributions from the student model}}
        \State $q^i=\mathcal{T}(\mathbf{x};\mathbf{y}^{i-1})$ \Comment{\textit{probability distributions from the teacher model}}
        \State Compute $\mathcal{L}_{kd}^i(p^i,q^i)$ based on Eq.\eqref{eq:new_kd_loss}
        \State $\mathcal{L}_{kd} \leftarrow \mathcal{L}_{kd} + \mathcal{L}_{kd}^i$
        \State $\mathbf{y}^i=\argmax(p^i)$ \Comment{\textit{student predictions as inputs in the next iteration}}
        \EndFor
        \State $\mathcal{L}_{word\raisebox{0mm}{-}kd} \leftarrow (1-\alpha)\mathcal{L}_{ce}+\frac{\alpha}{N}\mathcal{L}_{kd}$
    \end{algorithmic}
\end{algorithm}

\begin{table*}[]
    \centering
    \resizebox{\linewidth}{!}{
    \begin{tabular}{l|c|c|c|c|c|c}
        \bottomrule
        \multirow{2}{*}{\textbf{Methods}} & \multicolumn{2}{c|}{\textbf{WMT'14 En-De}} & \multicolumn{2}{c|}{\textbf{WMT'14 En-Fr}} & \multicolumn{2}{c}{\textbf{WMT'16 En-Ro}} \\
        \cline{2-7}
        & BLEU & COMET & BLEU & COMET & BLEU & COMET \\
        \hline
        \hline
        \textit{Student} (\textit{Transformer}$_{base}$) & 27.42$_{\pm 0.01}$ & 48.11$_{\pm 1.04}$ & 40.97$_{\pm 0.14}$ & 62.19$_{\pm 0.11}$ & 33.59$_{\pm 0.15}$ & 50.96$_{\pm 0.43}$ \\
        \quad + Word-KD \cite{seq_kd} & 28.03$_{\pm 0.10}$ & 51.59$_{\pm 0.23}$ & 41.10$_{\pm 0.11}$ & 63.81$_{\pm 0.14}$ & 33.77$_{\pm 0.01}$ & 53.15$_{\pm 0.26}$ \\
        \quad + Seq-KD \cite{seq_kd} & 28.22$_{\pm 0.02}$ & 51.23$_{\pm 0.15}$ & 41.44$_{\pm 0.02}$ & 63.12$_{\pm 0.14}$ & 33.69$_{\pm 0.02}$ & 50.63$_{\pm 0.11}$ \\
        \quad + BERT-KD \cite{chen-etal-2020-distilling}$^{\dagger}$ & 27.53 & - & - & - & - & - \\
        \quad + Seer Forcing \cite{feng-etal-2021-guiding} & 27.56$_{\pm 0.10}$ & 50.60$_{\pm 0.12}$ & 40.97$_{\pm 0.01}$ & 62.95$_{\pm 0.39}$ & 33.77$_{\pm 0.09}$ & 51.41$_{\pm 0.60}$ \\
        \quad + CBBGCA \cite{zhou-etal-2022-confidence}$^{\dagger}$ & 28.36 & - & 41.54 & - & - & - \\
        \quad + Annealing KD \cite{jafari-etal-2021-annealing} & 27.91$_{\pm 0.10}$ & 51.58$_{\pm 0.03}$ & 41.20$_{\pm 0.13}$ & 63.59$_{\pm 0.09}$ & 33.67$_{\pm 0.09}$ & 52.22$_{\pm 1.02}$ \\
        \quad + Selective-KD \cite{wang-etal-2021-selective} & 28.24$_{\pm 0.21}$ & 52.15$_{\pm 0.42}$ & 41.25$_{\pm 0.04}$ & 64.24$_{\pm 0.01}$ & 33.74$_{\pm 0.02}$ & 53.05$_{\pm 0.28}$ \\
        \quad + TIE-KD (ours) & \textbf{28.46$^{\ast}_{\pm 0.01}$} & \textbf{52.63$^{\ast}_{\pm 0.09}$} & \textbf{41.57$^{\ast}_{\pm 0.08}$} & \textbf{65.06$^{\ast}_{\pm 0.44}$} & \textbf{34.70$^{\ast}_{\pm 0.07}$} & \textbf{55.76$^{\ast}_{\pm 0.21}$} \\
        \hline
        \textit{Teacher} (\textit{Transformer}$_{big}$) & {\it 28.81} & \textit{53.20} & {\it 42.98} & \textit{69.58} & {\it 34.70} & \textit{57.04} \\
        \toprule
    \end{tabular}
    }
    \caption{BLEU scores (\%) and COMET \cite{rei-etal-2020-comet} scores (\%) on three translation tasks. Results with $^{\dagger}$ are taken from the original papers. Others are our re-implementation results using the released code with the same setting in Sec.\ref{sec:sec5_2} for a fair comparison. We report average results over 3 runs with random initialization. Results with $\ast$ are statistically \cite{koehn-2004-statistical} better than the vanilla Word-KD with $p<0.01$.}
    \label{tab:main_results}
\end{table*}
Given that the large overlap between the top-1 predictions and ground-truth targets limits the amount of additional knowledge from the teacher during word-level KD, introducing data without ground-truth targets for KD could be helpful to mitigate this issue.
Inspired by previous studies on decoder-side data manipulation \cite{zhang-etal-2019-bridging,teaforn,liu-etal-2021-confidence,liu-etal-2021-scheduled-sampling,xie2021target}, we design an iterative knowledge distillation procedure to expose more target-side data for KD.

Specifically, as shown in Algorithm \ref{algo:iter_kd}, at each training step, we conduct KD for $N$ iterations (line 3), by using the predictions of the student in the current iteration as the decoder-side inputs for KD in the next iteration (line 8). 
Generally, these predictions can be regarded as similar but new inputs compared to the original target inputs. 
Meanwhile, there is no ideal ground-truth target for these inputs since they are usually not well-formed sentences.
Then during each iteration, we collect the loss of KD according to Eq.\eqref{eq:new_kd_loss} (lines 4$\sim$7) and average it across all the iterations (line 10).
Since all the supervision signals are from the teacher after the first iteration, the knowledge of the teacher model will be more exploited during the following iterations and thus the potential of word-level KD can be more released. 

\section{Experiments} \label{sec:experiments}
\subsection{Datasets} \label{sec:datasets}
We conduct experiments on three commonly-used WMT tasks, {\em i.e.}, the WMT'14 English to German (En-De), WMT'14 English to French (En-Fr) and WMT'16 English to Romanian (En-Ro). 
For all these tasks, we share the source and the target vocabulary and segment words into subwords using byte pair encoding (BPE) \cite{sennrich-etal-2016-neural} with 32k merge operations. 
More statistics of the datasets can be found in Appendix \ref{sec:data_stat}.

\subsection{Implementation Details} \label{sec:sec5_2}
All our experiments are conducted based on the open-source toolkit fairseq \cite{ott2019fairseq} with FP16 training \cite{ott-etal-2018-fp16}. 
By default, we follow the big/base setting \cite{transformer} to implement the teacher/student models in our experiments.
More training and evaluation details can be referred to Appendix \ref{sec:train_detail}.
For word-level KD-based methods, we set the $\alpha$ in Eq.\eqref{eq:ori_final_loss} to 0.5 following \citet{seq_kd}.
For our method, we set top-$k$ in Sec.\ref{sec:hrl} to 5 and iteration time $N$ in Sec.\ref{sec:iter_kd} to 3 on all three tasks. 
The selection of top-$k$ and $N$ are shown in Appendix \ref{sec:hyper}.


\subsection{Main Results}
We compare our proposed method with existing KD techniques in NMT (the detailed description of these compared techniques can be referred to Appendix \ref{sec:systems}) on three WMT tasks.
To make the results more convincing, we report both BLEU and COMET \cite{rei-etal-2020-comet} scores in Tab.\ref{tab:main_results}.
Using Transformer$_{big}$ as the teacher, our method can boost the Transformer$_{base}$ students by +1.04/+0.60/+1.11 BLEU scores and +4.52/+2.57/+4.80 COMET scores on three tasks, respectively. 
Compared to the vanilla Word-KD baseline, our method can outperform it significantly on all translation tasks, which verifies the effectiveness of our proposed solutions.
Additionally, as a word-level KD method, our TIE-KD can outperform Seq-KD on all three tasks and even achieves fully competitive results with the teacher on En-Ro, which demonstrates that the potential of Word-KD can be largely released by our method. 

\section{Analysis}
\subsection{Ablation Study}

To separately verify the effectiveness of our solutions for the two issues in vanilla word-level KD, we conduct an ablation study on WMT'14 En-De task and record the results in Tab.\ref{tab:ablation_study}. 
When only adding hierarchical ranking loss to vanilla word-level KD, the BLEU scores and the TA rates gain by +0.3/+0.22 and +0.32/+0.47 on the validation/test set, respectively. 
It reflects that KL divergence only provides a loose constraint on the learning of the top-1 information from the teacher, while our hierarchical ranking loss helps to explicitly grasp this core information. 
When only using iterative KD, the student also improves by +0.36/+0.25 BLEU scores and +0.18/+0.28 TA rates. 
It indicates that our iterative KD can effectively release the potential of word-level KD by introducing data without ground-truth targets. 
When combined together, the two solutions finally compose our TIE-KD and can yield further improvement on both metrics. 
Therefore, the two issues in word-level KD are orthogonal and our proposed solutions are complementary to each other. 
\begin{table}[]
    \centering
    \resizebox{\linewidth}{!}{
    \begin{tabular}{l|c|c|c|c}
        \bottomrule
        \multirow{2}{*}{\textbf{Methods}} & \multicolumn{2}{c|}{\textbf{Validation Set}} & \multicolumn{2}{c}{\textbf{Test Set}} \\
        \cline{2-5}
        ~ & BLEU & TA & BLEU & TA \\
        \hline
        vanilla Word-KD & 26.66 & 88.98 & 28.03 & 88.46 \\
        \quad + $\mathcal{L}_{hr}$ & 26.96 & 89.30 & 28.25 & 88.93 \\
        \quad + iterative KD & 27.02 & 89.16 & 28.28 & 88.74 \\
        \quad + both (TIE-KD) & \textbf{27.13} & \textbf{89.50} & \textbf{28.46} & \textbf{89.11} \\
        \toprule
    \end{tabular}
    }
    \caption{Ablation study on the WMT'14 En-De task.}
    \label{tab:ablation_study}
\end{table}

\subsection{Combination With Sequence-Level KD} \label{sec:combine}
According to \cite{seq_kd}, word-level KD can be well combined with sequence-level KD and yields better performance. 
As a word-level KD approach, our TIE-KD can also theoretically be combined with sequence-level KD.
We verify this on the WMT'14 En-De task and list the results in Tab.\ref{tab:combine}.
Like Word-KD, our TIE-KD can also achieve better performance when combined with Seq-KD and is also better than ``Word-KD + Seq-KD'', indicating the superiority of our method and its high compatibility with sequence-level KD.
\begin{table}[t]
    \centering
    \begin{tabular}{l|c|c}
        \bottomrule
        \textbf{Methods} & \textbf{BLEU} & $\Delta$ \\
        \hline
        \hline
        Student (Transformer$_{base}$) & 27.42 & ref. \\
        \hline
        Word-KD & 28.03 & +0.61 \\
        Seq-KD & 28.22 & +0.80 \\
        TIE-KD & 28.46 & +1.04 \\
        \hline
        Word-KD + Seq-KD  & 28.48 & +1.06 \\
        TIE-KD + Seq-KD & \textbf{28.66} & \textbf{+1.24} \\
        \hline
        Teacher (Transformer$_{big}$) & 28.81 & +1.39 \\
        \toprule
    \end{tabular}
    \caption{Combination with sequence-level KD and word-level KD methods on the WMT'14 En-DE task.}
    \label{tab:combine}
\end{table}

\subsection{Can a Stronger Teacher Teach a Better Student in NMT?}
\begin{figure}
    \centering
    \includegraphics[width=\linewidth]{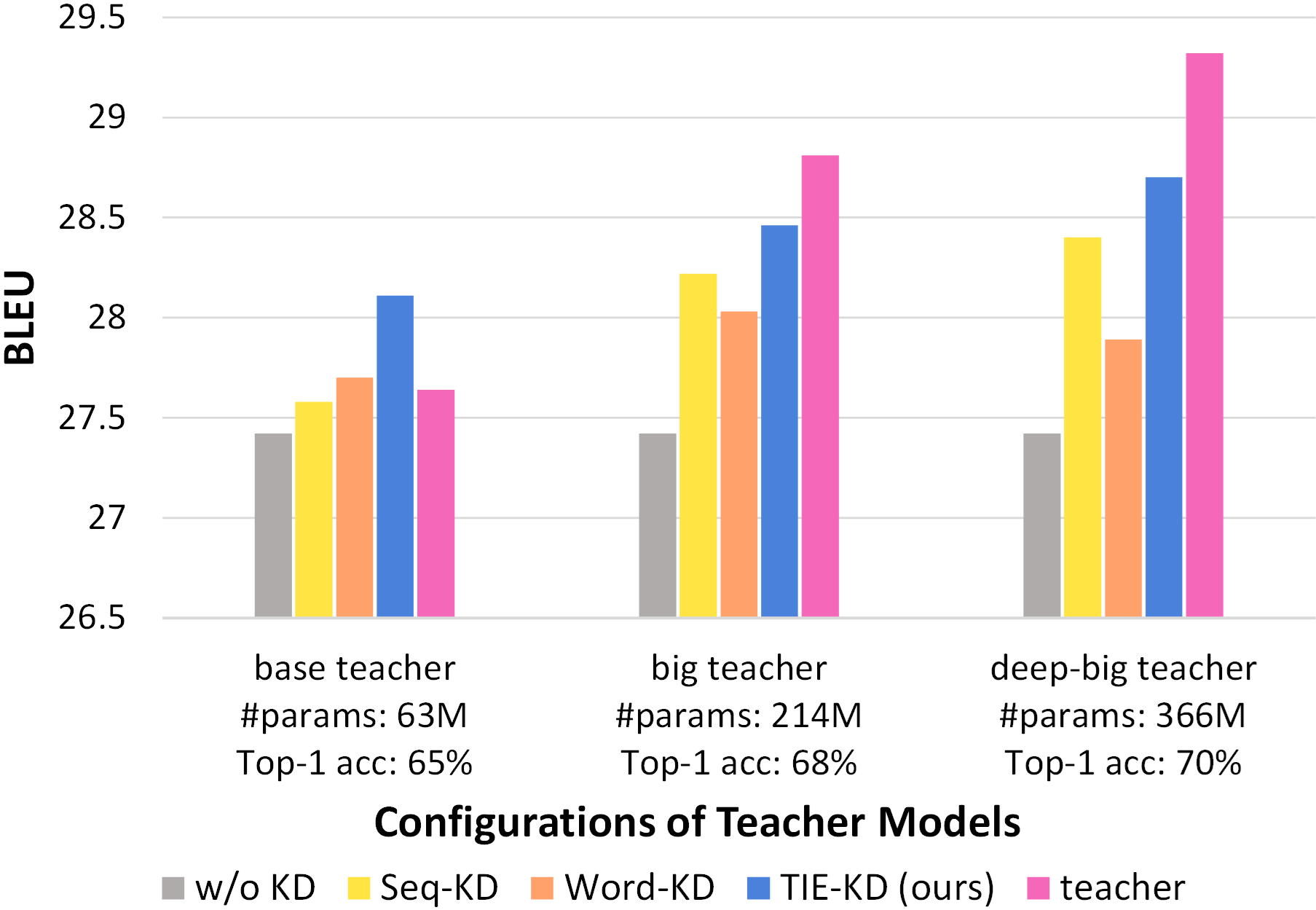}
    \caption{Performance of KD techniques with different teacher models on the test set of the WMT'14 En-De task.}
    \label{fig:diff_teacher}
\end{figure}
Among the prior literature on KD \cite{kd_efficacy,jin2019knowledge,takd,guo2020reducing,jafari-etal-2021-annealing,qiu2022better}, a general consensus is that a large teacher-student capacity gap may harm the quality of KD.
We also check this problem in NMT by using teachers of three model sizes.
Besides the default configuration (\textit{i.e.}, Transformer$_{big}$) in our experiments above, we also add Transformer$_{base}$ setting as the weaker teacher and Transformer$_{deep\raisebox{0mm}{-}big}$ setting with 18 encoder layers and 6 decoder layers as the stronger teacher\footnote{To stably train a deeper Transformer, we use Admin \cite{liu-etal-2020-understanding} in layer normalization.}.
We compare our method with word- and sequence-level KD under these teachers in Fig.\ref{fig:diff_teacher} and draw several conclusions:
\begin{enumerate}[(1)]
    \item The stronger teacher can bring improvement to sequence-level KD but fails to word-level KD, where the reason may be the less additional knowledge from the stronger teacher due to its higher top-1 accuracy (68\%$\rightarrow$70\%).
    \item As a word-level KD method, our TIE-KD instead brings conspicuous improvement with the stronger teacher, indicating that our method can exploit more knowledge from the teacher.
    \item Under the weaker teacher, the student from our method even significantly surpasses the teacher, while other methods are largely limited by the performance of the teacher, demonstrating the high generalizability of our TIE-KD to different teacher-student capacity gaps.
\end{enumerate}

\subsection{Why is the Top-1 Information Important in KD?}
The decoding process of language generation models can be regarded as a sequential decision-making process \cite{yu2017seqgan,arora-etal-2022-exposure}. 
As mentioned in Sec.\ref{sec:expand_seq}, during decoding, beam search tends to pick the top-1 predictions of the NMT model on each beam and finally selects the most probable beam. 
Thus, the top-1 information (including both the top-1 word index and its corresponding probability) of the teacher model largely represents its decision on each decoding step, which is exactly what we expect the student model to learn from the teacher through KD in NMT.
Therefore, the top-1 information can be seen as the embodiment of the knowledge of the teacher model in NMT tasks and should be emphatically learned by the student models.

\section{Related Work} \label{sec:related_work}
\citet{seq_kd} first introduce word-level KD for NMT and further propose sequence-level KD for better performance.
Afterward, \citet{wang-etal-2021-selective} investigate the effectiveness of different types of tokens in KD and propose selective KD strategies.
Moreover, \citet{wu2020skip} distill the internal hidden states of the teacher models into the students and also obtain promising results.
In the field of non-autoregressive machine translation (NAT), KD from autoregressive models has become a $\textit{de facto}$ standard to improve the performance of NAT models \cite{gu2017non,zhou2019understanding,gu2019levenshtein}.
Also, KD has been used to enhance the performance of multilingual NMT \cite{tan2019multilingual,sun-etal-2020-knowledge-distillation}.
Besides, similar ideas can be found when introducing external information to NMT models.
For example, \citet{baziotis-etal-2020-language} use language models as teachers for low-resource NMT models.
\citet{chen-etal-2020-distilling} distill the knowledge from fine-tuned BERT into NMT models.
\citet{feng-etal-2021-guiding} and \citet{zhou-etal-2022-confidence} leverage KD to introduce future information to the teacher-forcing training of NMT models.

Differently, in this work, 1) we aim to explore where the knowledge hides in KD and unveil that it comes from the top-1 information of the teacher and further improve KD from this perspective; 2) we try to build a connection between two kinds of KD techniques in NMT and reveal their common essence, providing new directions for future work.

\section{Conclusion}
In this paper, we explore where the knowledge hides in KD for NMT and unveil that it comes from the top-1 information of the teacher.
This finding reflects the connection between word- and sequence-level KD and reveals the common essence of both KD techniques in NMT.
From this perspective, we further propose a top-1 information enhanced knowledge distillation (TIE-KD) to address the two issues in vanilla word-level KD.
Experiments on three WMT tasks prove the effectiveness of our method.
Besides, we investigate the performance of the existing KD techniques in NMT and our method under different teacher-student capacity gaps and show the stronger generalizability of our method on various gaps.

\section*{Limitations}
Although our method has achieved outstanding performance compared to current KD techniques, it is still a word-level KD method and also suffers from some limitations in vanilla word-level KD, \textit{e.g.}, the \textit{exposure bias} as analyzed in Appendix \ref{sec:theoretical}.
How to design a unified and more powerful KD method from the perspective of the connection between word- and sequence-level KD still remains unsolved.
We will leave this for the future work.
Moreover, our study focuses on the mainstream KD techniques in NMT, which transfer knowledge through teachers' predictions, while some other KD techniques, like directly distilling the hidden states \cite{wu2020skip}, are not within the scope of this study and thus not included.

\section*{Acknowledgements}
The research work described in this paper has been supported by the National Key R\&D Program of China (2020AAA0108001) and the National Nature Science Foundation of China (No. 61976016 and 61976015). Wenjuan Han is supported by the Talent Fund of Beijing Jiaotong University (2023XKRC006). The authors would like to thank the anonymous reviewers for their valuable comments and suggestions to improve this paper.

\bibliography{anthology,custom}
\bibliographystyle{acl_natbib}

\clearpage
\newpage
\appendix

\section{A Theoretical Analysis on the Connection Between Word- and Sequence-level KD} \label{sec:theoretical}
We can directly consider the KL divergence loss of word-level KD in Eq.\eqref{eq:kl_loss} as its training objective and convert it into the equivalent form of the cross-entropy loss. For simplicity, we omit the $\theta_t$ in $q(\cdot)$ and $\theta_s$ in $p(\cdot)$ in following formulas:
\begin{align} \label{eq:kl2ce}
    \mathcal{L}^{word}_{kd}&=\!\sum_{j=1}^N\! D_{\rm KL}\Big(q(y_j|\mathbf{y}_{<j},\mathbf{x})\big|\big|p(y_j|\mathbf{y}_{<j},\mathbf{x})\Big) \nonumber \\
    &\Leftrightarrow-\sum_{j=1}^N\sum_{k\in\mathcal{V}}q(y_j=k|\mathbf{y}_{<j},\mathbf{x})\times \nonumber\\
    &\qquad\qquad\log p(y_j=k|\mathbf{y}_{<j},\mathbf{x}),
\end{align}
where $\mathcal{V}$ denotes the whole target-side vocabulary.
Then we can further separate the cross-entropy loss into the loss on the top-1 prediction $y_j^{t1}$ and the losses on other candidates in the vocabulary:
\begin{align} \label{eq:seperate_ce}
    \mathcal{L}^{word}_{kd}&=-\sum_{j=1}^N\sum_{k\in\mathcal{V}}q(y_j=k|\mathbf{y}_{<j},\mathbf{x})\times \nonumber\\
    &\qquad\qquad\log p(y_j=k|\mathbf{y}_{<j},\mathbf{x}) \nonumber \\
    &=-\sum_{j=1}^N\Big(q(y_j^{t_1}|\mathbf{y}_{<j},\mathbf{x})\log p(y_j^{t_1}|\mathbf{y}_{<j},\mathbf{x}) \nonumber \\
    &\quad+\sum_{k\in\mathcal{V}\backslash\{ y_j^{t_1}\}}q(y_j=k|\mathbf{y}_{<j},\mathbf{x})\times \nonumber\\
    &\qquad\qquad\log p(y_j=k|\mathbf{y}_{<j},\mathbf{x})\Big) \nonumber \\
    &=-\sum_{j=1}^N\Big(q(y_j^{t_1}|\mathbf{y}_{<j},\mathbf{x})\log p(y_j^{t_1}|\mathbf{y}_{<j},\mathbf{x}) \nonumber \\
    &\qquad\qquad +R(y_j^{t_1})\Big),
\end{align}
where $R(y_j^{t_1})$ represents the cross-entropy loss on the remaining candidates except for the top-1 prediction $y_j^{t_1}$ and can be regarded as a regularization term for the former one.
As empirically verified in Sec.\ref{sec:sec3}, we can do the following approximation by omitting $R(y_j^{t_1})$ in Eq.\eqref{eq:seperate_ce}:
\begin{align} \label{eq:approx_word}
    \mathcal{L}^{word}_{kd}&=-\sum_{j=1}^N\Big(q(y_j^{t_1}|\mathbf{y}_{<j},\mathbf{x})\log p(y_j^{t_1}|\mathbf{y}_{<j},\mathbf{x}) \nonumber \\
    &\qquad\qquad +R(y_j^{t_1})\Big) \nonumber \\
    &\approx-\sum_{j=1}^N q(y_j^{t_1}|\mathbf{y}_{<j},\mathbf{x})\log p(y_j^{t_1}|\mathbf{y}_{<j},\mathbf{x}).
\end{align}
Thus, we obtain the approximate form of the training objective of word-level KD.

Now we consider the training objective of sequence-level KD in Eq.\eqref{eq:decompose}. According to the results in Sec.\ref{sec:expand_seq}, we can also assume that optimizing using all targets is approximately equal to optimizing using top-1 targets:
\begin{align} \label{eq:approx_seq}
    \mathcal{L}^{seq}_{kd}&=-\sum^N_{j=1} \log p(\widehat{y}_{j}|\widehat{\mathbf{y}}_{<j},\mathbf{x}) \nonumber \\
    &\approx-\sum^N_{j=1}\mathbbm{1}\{\widehat{y}_j=y_j^{t_1}\} \log p(y_j^{t_1}|\widehat{\mathbf{y}}_{<j},\mathbf{x}),
\end{align}
where $\mathbbm{1}\{\cdot\}$ is an indicator function.

Lastly, if we replace the different weight functions before the $\log(\cdot)$ function in Eq.\eqref{eq:approx_word} and Eq.\eqref{eq:approx_seq} with one function $f(\cdot)$:
\begin{equation} \label{eq:weight_func}
        f(j)=\left\{
        \begin{array}{ll}
            q(y_j^{t_1}|\mathbf{y}_{<j},\mathbf{x}), & { word\raisebox{0mm}{-}level} \\
            \\
            \mathbbm{1}\{\widehat{y}_j=y_j^{t_1}\}, & { sequence\raisebox{0mm}{-}level},
        \end{array}
        \right.\nonumber
\end{equation}
then we can derive a unified form of the objective for these two kinds of KD techniques:
\begin{equation}
    \mathcal{L}^{uni}_{kd}=-\sum_{j=1}^N f(j)\log p(y_j^{t_1}|\widetilde{\mathbf{y}}_{<j},\mathbf{x}),
\end{equation}
where $\widetilde{\mathbf{y}}_{<j}$ is the golden context $\mathbf{y}_{<j}$ in word-level KD and the model-generated context $\widehat{\mathbf{y}}_{<j}$ in sequence-level KD.

In this unified form, the only two variables are the weight function $f(\cdot)$ and the target-side previous context $\widetilde{\mathbf{y}}_{<j}$ in the condition of the probability $p(\cdot)$.
From this expression, it is clear that student models are encouraged to learn the top-1 predictions of the teacher to obtain teachers' knowledge at each time step in both KD techniques.
Therefore, we claim that the working mechanisms behind the two kinds of KD techniques are the same to some extent, although they look quite distinct on the surface.

Notably, we also conjecture that the context difference may explain why sequence-level KD generally outperforms word-level KD.
Autoregressive models trained with teacher-forcing suffer from \textit{exposure bias} due to the gap between the golden context in training and the model-generated context in inference \cite{ss,zhang-etal-2019-bridging}.
According to the above analysis, the same thing also happens in word-level KD.
However, sequence-level KD circumvents this problem by conditioning on model-generated contexts during distillation, thus leaving no gap between training and inference.
This conjecture can also be verified by the performance of sequence-level KD on WMT'16 En-Ro, where the teacher's translations achieve considerably high similarities (BLEU score $>$ 62) with the original target sentences, and the improvement brought by sequence-level KD is much less than the one on other datasets since the model-generated context is too close to the golden context.

\section{Why Not Re-normalize the Soft Target in ``\textit{w/o} correlation info''?} \label{sec:whynot_renorm}
We would like to explain this from the perspective of the loss function. As we analyzed in Eq.\eqref{eq:seperate_ce}, the loss of vanilla word-level KD is:
\begin{align} \label{eq:word_kd_ce_loss}
    \mathcal{L}^{word}_{kd}&=-\sum_{j=1}^N\Big(q(y_j^{t_1}|\mathbf{y}_{<j},\mathbf{x})\log p(y_j^{t_1}|\mathbf{y}_{<j},\mathbf{x}) \nonumber \\
    &\quad+\sum_{k\in\mathcal{V}\backslash\{ y_j^{t_1}\}}q(y_j=k|\mathbf{y}_{<j},\mathbf{x})\times \nonumber\\
    &\qquad\qquad\log p(y_j=k|\mathbf{y}_{<j},\mathbf{x})\Big).
\end{align}
Based on this, we remove all other probabilities in the soft target of the teacher except for the top-1 one to remove the ``correlation information", \textit{i.e.}, the second term of the loss in Eq.\eqref{eq:word_kd_ce_loss} is discarded:
\begin{align}
    \mathcal{L}^{nocorr}_{kd}=-\sum_{j=1}^N q(y_j^{t_1}|\mathbf{y}_{<j},\mathbf{x})\log p(y_j^{t_1}|\mathbf{y}_{<j},\mathbf{x}) \nonumber.
\end{align}
In this objective, the effect of KD is fully dominated by the top-1 information of the teacher.
If we try to re-normalize the soft target with an additional uniform distribution, the result of KD will be affected by the regularization term of this uniform distribution:
\begin{align} \label{eq:nocorr_kd_loss_with_reg}
    \mathcal{L}^{nocorr}_{kd}&=-\sum_{j=1}^N\Big(q(y_j^{t_1}|\mathbf{y}_{<j},\mathbf{x})\log p(y_j^{t_1}|\mathbf{y}_{<j},\mathbf{x}) \nonumber \\
    &\quad+\underbrace{u\sum_{k\in\mathcal{V}\backslash\{ y_j^{t_1}\}}\log p(y_j=k|\mathbf{y}_{<j},\mathbf{x}}_{uniform\ regularization})\Big)\nonumber,
\end{align}
where $u=\frac{1-q(y_j^{t_1}|\mathbf{y}_{<j},\mathbf{x})}{|\mathcal{V}|-1}$.
Another way to re-normalize the distribution is to directly let $q(y_j^{t_1}|\mathbf{y}_{<j},\mathbf{x})$ as 1, but the original top-1 probability information from the teacher will be lost.
Therefore, we keep the modified soft target in ``\textit{w/o} correlation info" unnormalized.

\section{Experimental Details}
\label{sec:appendix}

\subsection{Statistics of the Datasets} \label{sec:data_stat}
For the En-De task, the training data contains nearly 4.5M sentence pairs. 
We choose \textit{newstest2013} and \textit{newstest2014} as the validation set and the test set, respectively.
For the En-Fr task, there totally remains 35.8M sentence pairs after the cleaning procedure.
Then we choose \textit{newstest2013} and \textit{newstest2014} as the validation set and the test set, respectively.
For the En-Ro task, we directly use the pre-processed data from \citet{mehta2020delight} and there are about 608K sentence pairs in the training data.
Then \textit{newsdev2016} is selected as the validation set and \textit{newstest2016} is the test set.
The overall statistics of the datasets are listed in Table \ref{tab:dataset}.
\begin{table}[H]
    \centering
    \resizebox{\linewidth}{!}{
    \begin{tabular}{l|c|c|c|c}
        \bottomrule
        \textbf{Dataset} & \textbf{\#Train} & \textbf{\#Valid} & \textbf{\#Test} & \textbf{Vocab} \\
        \hline
        WMT'14 En-De & 4.5M & 3000 & 3003 & 37184 \\
        WMT'14 En-Fr & 35.8M & 3000 & 3003 & 36528 \\
        WMT'16 En-Ro & 608K & 1999 & 1999 & 34976 \\
        \toprule
    \end{tabular}
    }
    \caption{Statistics of the datasets for three WMT tasks.}
    \label{tab:dataset}
\end{table}
\subsection{Implementation Details and Model Configurations} \label{sec:train_detail}
\begin{table*}[]
    \centering
    \resizebox{\linewidth}{!}{
    \begin{tabular}{l|c|c|c|c|c|c}
        \bottomrule
        \multirow{2}{*}{\textbf{Hyperparameters}} & \multicolumn{2}{c|}{\textbf{WMT'14 En-De}} & \multicolumn{2}{c|}{\textbf{WMT'14 En-Fr}} & \multicolumn{2}{c}{\textbf{WMT'16 En-Ro}} \\
        \cline{2-7}
        & Student & Teacher & Student & Teacher & Student & Teacher \\
        \hline
        Embedding Dim & 512 & 1024 & 512 & 1024 & 512 & 1024 \\
        FFN Dim & 2048 & 4096 & 2048 & 4096 & 2048 & 4096 \\
        Encoder Layers & 6 & 6 & 6 & 6 & 6 & 6 \\
        Decoder Layers & 6 & 6 & 6 & 6 & 6 & 6 \\
        Attention Heads & 8 & 16 & 8 & 16 & 8 & 16 \\
        Residual Dropout & 0.1 & 0.3 & 0.1 & 0.3 & 0.1 & 0.3 \\
        Attention Dropout & 0.1 & 0.1 & 0.1 & 0.1 & 0.1 & 0.1 \\
        Activation Dropout & 0.1 & 0.1 & 0.1 & 0.1 & 0.1 & 0.1 \\
        Label Smoothing & 0.1 & 0.1 & 0.1 & 0.1 & 0.1 & 0.3 \\
        Learning Rate & 7e-4 & 5e-4 & 7e-4 & 5e-4 & 7e-4 & 5e-4 \\
        Learning Rate Decay & inverse sqrt & inverse sqrt & inverse sqrt & inverse sqrt & inverse sqrt & inverse sqrt \\
        Warmup Steps & 4000 & 4000 & 4000 & 4000 & 4000 & 4000 \\
        Layer Normalization & PostNorm & PostNorm & PostNorm & PostNorm & PostNorm & PostNorm \\
        Model Parameters & 63.2M & 214.4M & 62.8M & 213.8M & 62.0M & 212.2M \\
        Training Steps & 200K & 300K & 200K & 300K & 20 epochs & 30 epochs \\
        \toprule
    \end{tabular}
    }
    \caption{Training hyperparameters and model configurations of our experiments.}
    \label{tab:train_detail}
\end{table*}
\paragraph{Training.} To assure the reproducibility of our experimental results, we provide comprehensive training details and model configurations of our experiments in Tab.\ref{tab:train_detail}. 
All our experiments are conducted on 4 NVIDIA RTX 3090 GPUs with gradient accumulation step 2, and each batch on each GPU contains approximately 4096 tokens. 
We use Adam optimizer \cite{adam} with 4000 warmup steps to optimize models. 
To obtain strong teachers and enlarge the gaps between teacher models and student models, we train teachers for 50\% more steps than the corresponding students.
Then we use the checkpoint with the highest BLEU of the teacher on the validation set to conduct distillation.
\paragraph{Evaluation.}During inference, we set beam size to 4 and length penalty to 0.6 for En-De and En-Fr. 
For En-Ro, we set beam size to 5 and length penalty to 1.2. 
For a more convincing evaluation, we use \textit{multibleu.perl} to calculate case-sensitive BLEU and \textit{unlabel-comet}\footnote{\url{https://github.com/Unbabel/COMET}} to calculate COMET scores \cite{rei-etal-2020-comet} for all three tasks. 
For student models, we average the last 5 checkpoints for evaluation following \citet{transformer}.
We use the paired bootstrap resampling methods \cite{koehn-2004-statistical} for the statistical significance test.
For the En-De task and the En-Fr task, we evaluate and save the checkpoint every 5000 training steps.
For the En-Ro task, since the models tend to overfit, we only train students for 20 epochs and save the checkpoint after every epoch.

\subsection{Compared Systems and Hyperparameters} \label{sec:systems}
\paragraph{Transformer.} We follow the standard base/big model configurations \cite{transformer} to implement the student/teacher models.
\paragraph{Word-KD.} The standard method to conduct word-level KD in NMT proposed by \citet{seq_kd}.
\paragraph{Seq-KD.} \citet{seq_kd} also propose a sequence-level KD approach that directly substitutes the original target-side training data with the translations of the teacher from beam search.
In our experiments, the hyperparameters of beam search keep the same with the inference stage. 
\paragraph{BERT-KD.} \citet{chen-etal-2020-distilling} propose to distill the knowledge from BERT \cite{bert} for text generation tasks.
\paragraph{Seer Forcing.} \citet{feng-etal-2021-guiding} design a seer forcing method for NMT to distill future information to the teacher forcing.
Following the suggestion in \cite{feng-etal-2021-guiding}, we set the $\alpha$ in their paper to 0.5 for both En-De and En-Fr, and 0.25 for En-Ro.
Besides, we set the seer dropout to 0.1 for En-De and En-Fr and 0.2 for En-Ro. 
\paragraph{CBBGCA.} \citet{zhou-etal-2022-confidence} also propose to distill bi-directional contextual information in CMLM for uni-directional training of NMT based on the confidence of the NMT model.
\paragraph{Annealing KD.} Our implementation of the method in \cite{jafari-etal-2021-annealing} which gradually anneals the temperature of the teacher during KD.
Different from the original paper, we use the KL divergence as the loss function of KD instead of Mean Squared Error (MSE) due to its better performance on NMT tasks.
In our carefully chosen recipe, we set the max temperature to 1.1 and gradually reduce it to 1.0 during the first 2/3 epochs.
Then we use vanilla CE loss to train the student model for the remaining 1/3 epochs.
\paragraph{Selective-KD.} \citet{wang-etal-2021-selective} investigate the effectiveness of different data for distillation and propose a knowledge selection method for selecting more valuable data for word-level KD.
In our experiments, we choose the global-level selection that performs better according to \citet{wang-etal-2021-selective}.

\section{Hyperparameter Selection} \label{sec:hyper}
\subsection{Effect of Hierarchical Ranking Range $\bm{k}$}
\label{sec:effect_topk}
\begin{figure}[H]
    \centering
    \includegraphics[width=\linewidth]{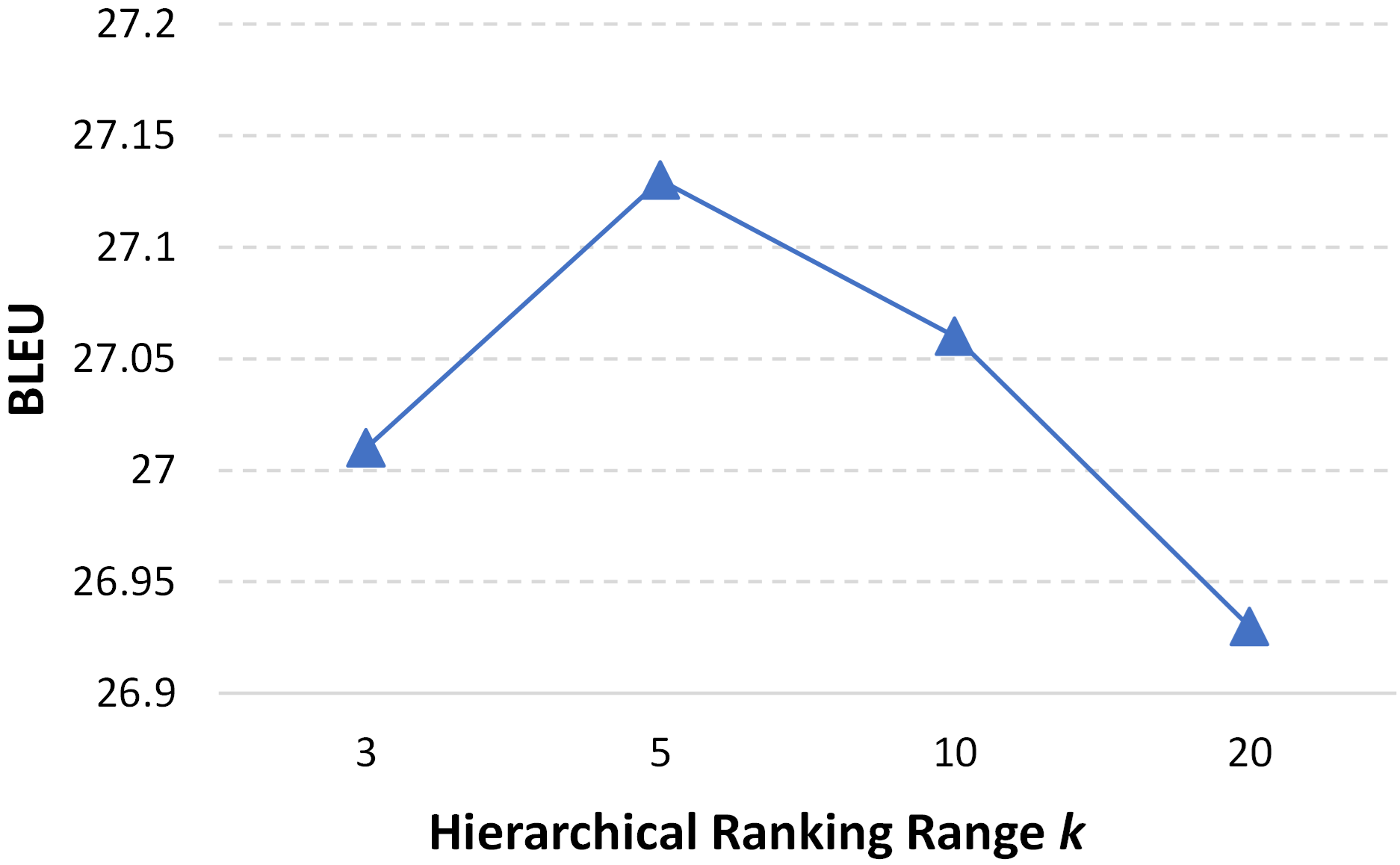}
    \caption{BLEU scores of our method with different $k$ on the validation set of the WMT'14 En-De task.}
    \label{fig:hyper_k}
\end{figure}

In this section, we investigate the effect of $k$ in hierarchical ranking loss on our method.
We search $k$ in [3, 5, 10, 20] and compare their performance on the validation set of the WMT'14 En-De task.
As shown in Fig.\ref{fig:hyper_k}, our method performs best when $k$ is set to 5. 
Thus, we keep $k$ to 5 for all three tasks in our experiments.
\subsection{Effect of Iteration Times $\bm{N}$}
\label{sec:effect_iter_time}
\begin{figure}[H]
    \centering
    \includegraphics[width=\linewidth]{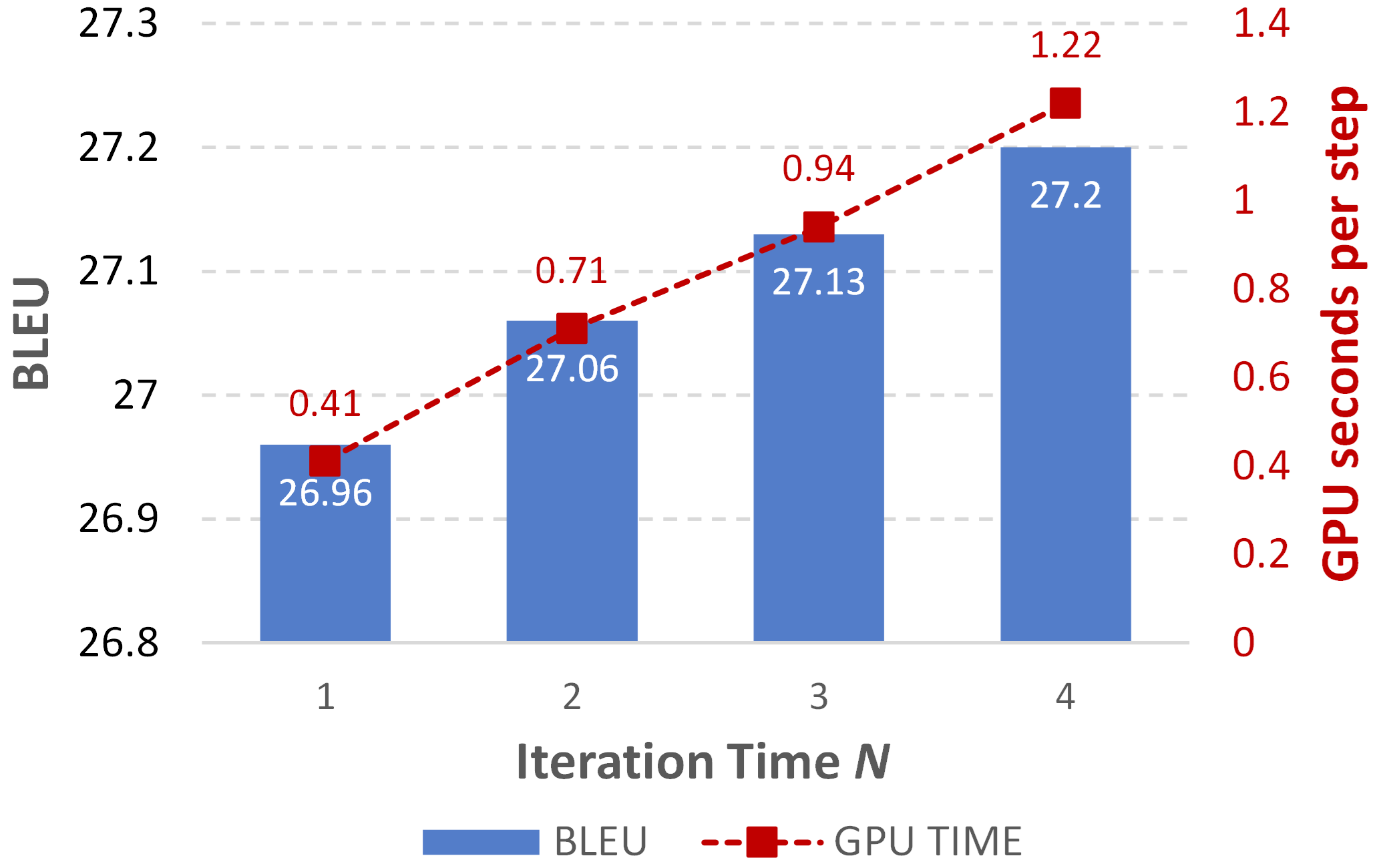}
    \caption{BLEU scores of our method with different iteration times $N$ on the validation set of the WMT'14 En-De task and the corresponding training costs.}
    \label{fig:hyper_N}
\end{figure}

Since our method includes several iterations of KD, we further investigate the effects of the iteration times on the performance of our method.
Intuitively, with more iteration times, more knowledge will be exploited from the teacher, while the computational cost will also increase.
To check this, we try each iteration time in [1, 2, 3, 4] and record the corresponding performance and training time in Fig.\ref{fig:hyper_N}.
It is obvious that the performance of our method gradually improves with $N$ increasing, while the training time per step also linearly increases.
Balancing the cost and the performance, we choose 3 as the final iteration time.

\end{document}